\documentclass{article}
\usepackage{spconf,amsmath,graphicx}
\usepackage{booktabs}
\usepackage{amsfonts}

\title{Syntactic Representation Learning for Neural Network based TTS with syntactic parse Tree Traversal}
\name{Changhe Song$^1$, Jingbei Li$^1$, Yixuan Zhou$^1$, Zhiyong Wu$^{1, *}$ \thanks{* Corresponding author.}, Helen Meng$^{2}$}
\address{
    $^1$ Shenzhen International Graduate School, Tsinghua University, Shenzhen, China\\
    $^2$ Department of Systems Engineering and Engineering Management, \\
         The Chinese University of Hong Kong, Hong Kong SAR, China\\
    \small{
        \{sch19, lijb19, zhouyx20\}$@$mails.tsinghua.edu.cn, 
        \{zywu, hmmeng\}$@$se.cuhk.edu.hk
    }
}

\begin{document}
\ninept
\maketitle

\begin{abstract}
Syntactic structure of a sentence text is correlated with the prosodic structure of the speech that is crucial for improving the prosody and naturalness of a text-to-speech (TTS) system. 
Nowadays TTS systems usually try to incorporate syntactic structure information with manually designed features based on expert knowledge. 
In this paper, we propose a syntactic representation learning method based on syntactic parse tree traversal to automatically utilize the syntactic structure information. 
Two constituent label sequences are linearized through left-first and right-first traversals from constituent parse tree.
Syntactic representations are then extracted at word level from each constituent label sequence by a corresponding uni-directional gated recurrent unit (GRU) network. 
Meanwhile, nuclear-norm maximization loss is introduced to enhance the discriminability and diversity of the embeddings of constituent labels. 
Upsampled syntactic representations and phoneme embeddings are concatenated to serve as the encoder input of Tacotron2. 
Experimental results demonstrate the effectiveness of our proposed approach, with mean opinion score (MOS) increasing from $3.70$ to $3.82$ and ABX preference exceeding by $17\%$ compared with the baseline. In addition, for sentences with multiple syntactic parse trees, prosodic differences can be clearly perceived from the synthesized speeches.

\end{abstract}


\begin{figure*}[!htb]
	\centering
	\includegraphics[width=0.95\linewidth]{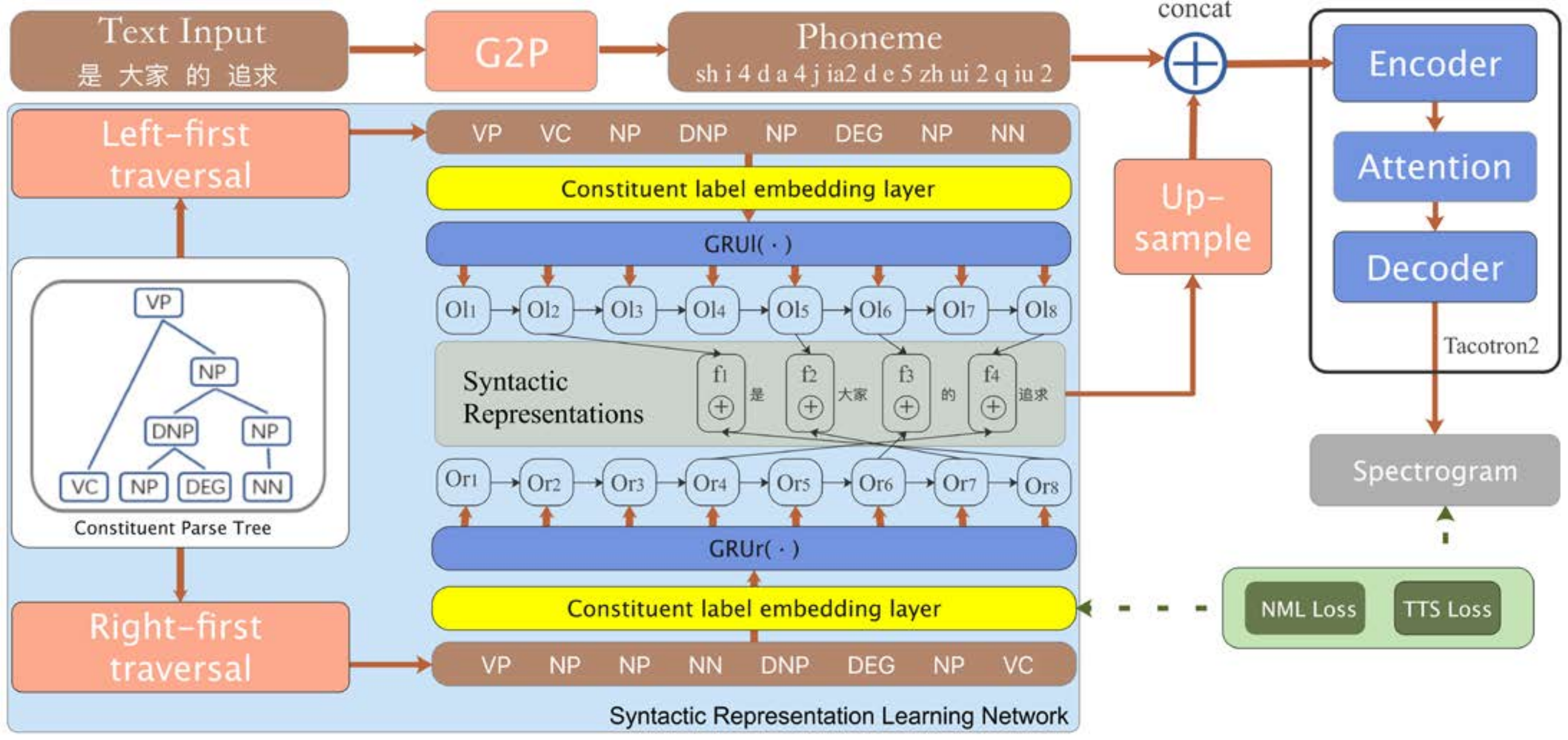}
	\caption{The structure of proposed syntactic representation learning network.}
	\label{fig:model_structure}
\end{figure*}

\section{Introduction}
\label{sec:intro}

Recently neural network based text-to-speech (TTS) systems have achieved certain success in prosody and naturalness of synthesized speech over conventional methods \cite{sotelo2017char2wav, wang2017tacotron, ping2017deep, shen2018natural}. 
By applying encoder-decoder framework with attention \cite{cho2014learning}, these systems can directly predict speech parameters from graphemes or phonemes by learning acoustic and prosodic patterns via a flexible mapping from linguistic to acoustic space. 
However, the learnt prosodic patterns only contain part of prosodic structural information \cite{shen2018natural}, resulting in poor prosody and naturalness performance even improper prosody.

To further improve prosody and naturalness of synthesized speech, adding prosodic structure annotations such as tones and break indices (ToBI) labels \cite{silverman1992tobi} or other prosodic structure labels \cite{poesio1999mate} to the input sequence of neural network based TTS models has been proposed.
Prosodic structure annotations need to be subjectively labeled from speech, which is time-consuming.

Although these annotations can be automatically annotated by training another prosodic structure prediction model \cite{rosenberg2010autobi}, the accuracy of predicted prosodic structure labels is still limited by using subjectively labeled annotations as the ground-truths.
The high correlation between syntactic structure and prosodic information has been proved by successful syntactic-to-prosodic mapping \cite{zhang2012break, che2014improving}.
A set of rule-based syntactic features such as part-of-speech (POS) and positions of the current word in parent phrases are proposed and used in hidden Markov model (HMM) based acoustic model \cite{dall2016redefining}.
To utilize more syntactic structure information, phrase structure based feature (PSF) and word relation based feature (WRF) are proposed in neural network based TTS \cite{guo2019exploiting}.
PSF and WRF expand the set of syntactic features used in HMM model.
More features such as highest-level phrase beginning with current word (HBCW) and lowest common ancestor (LCA) are further introduced to model syntactic structure \cite{guo2019exploiting}.

However, the expanded features are still manually designed features rather than automatically learned high-level representations.
PSF only contains features from limited layers of the whole syntactic tree structure.
WRF only exposes the information of partial nodes and edges from the whole syntactic parse tree.

To maker better use of the syntactic information, motivated by the syntactic parse tree traversal approach in neural machine translation \cite{li2017modeling}, we propose a syntactic representation learning method to further improve the prosody and naturalness of synthesized speech in neural network based TTS. 
Syntactic parse tree is linearized into two constituent label sequences through left-first and right-first traversal.
Then syntactic representations are extracted from the constituent label sequences using different uni-directional GRU network for each sequence.
After which, the syntactic representations are up-sampled from word level to phoneme level and concatenated with phoneme embeddings. 
Tacotron 2 is employed to generate spectrogram from the concatenated syntactic representations and phoneme embeddings, with Griffin-Lim \cite{griffin1984signal} to reconstruct the waveform.
Nuclear-norm maximization loss (NML) is introduced to the constituent label embedding layer to enhance discriminability and diversity. 
Compared to only hiring left-first traversal \cite{li2017modeling}, right-first traversal is proposed to alleviate the ambiguity.

Experimental results show that our proposed model outperforms the baseline in terms of prosody and naturalness.
Mean opinion score (MOS) increases from $3.70$ to $3.82$ compared with the baseline approach (t-test, p=0.0079).
ABX preference rate exceeds the baseline approach by $17\%$.
For sentences with multiple different syntactic parse trees, prosodic differences can be clearly perceived from corresponding synthesized speeches.


\section{Methodology}
\label{sec:metho}

Fig.\ref{fig:model_structure} shows the framework of our proposed method.
Our work mainly focuses on introducing a trainable syntactic structure information extractor as part of neural network based TTS system to improve the prosody and naturalness of the synthesized speech.

\subsection{Syntactic representation learning}
\label{ssec:metho_3_proposed}

To provide high-level syntactic representations with rich syntactic information to neural network based TTS system, we propose a syntactic representation learning network based on syntactic parse tree traversal.
Constituent parse trees are extracted including labels and tree structure of constituents.
To represent the tree structure for neural network based TTS, depth-first traversals are of possible to use to linearize the syntactic parse tree to constituent sequences.
Since any single tree traversal algorithm will map multiple syntactic parse trees to a same sequence, both left-first and right-first are proposed to use to alleviate the ambiguity.
The sequence of constituent labels generated by the two traversals can be formulated as the following equations:
\begin{equation}
\begin{split}
\small
\label{eq:clcr_1}
C_l &= [c_l^1, ..., c_l^m] \\
C_r &= [c_r^1, ..., c_r^m]
\end{split}
\end{equation}
where $C_l$ and $C_r$ are the constituent label sequences generated from the left-first and right-first traversals respectively, $c_l^i$ and $c_r^i$ are constituent labels, $m$ is the length of sequences.
The constituent labels are then embedded by a shared embedding layer and modeled by two different uni-directional GRU networks, one GRU network for each sequence.  
The process can be represented as: 
\begin{equation}
\begin{split}
\small
\label{eq:clcr_2}
&\hat{C}_l, \hat{C}_r = Embedding(C_l, C_r) \\
&O_l = GRU_l( \hat{C}_l ) \\
&O_r = GRU_r( \hat{C}_r ) 
\end{split}
\end{equation}
where $\hat{C_l}$ and $\hat{C_r}$ are embedding sequences of constituent labels, $GRU_l(\cdot)$ and $GRU_r(\cdot)$ are two different uni-directional GRUs, $O_l$ and $O_r$ are the outputs of $GRU_l(\cdot)$ and $GRU_r(\cdot)$ respectively.

Syntactic features are the concatenations of the outputs of GRUs for each word, which can be formulated as:
\begin{equation}
\small
\label{eq:clcr_3}
f_i = [O_l^{p_l^i},  O_r^{p_r^i}], i \in \{1, 2, ..., w\}
\end{equation}
where $p_l^i$ and $p_r^i$ are the positions of the $i$-th word in $C_l$ and $C_r$, $w$ is the number of words of the input text, $f_i$ is the learnt syntactic representation.

\subsection{Nuclear-norm Maximization Loss}
\label{ssec:metho_4_nml}

To improve the discriminability and diversity of the embeddings of the syntactic labels, global nuclear-norm maximization loss (NML) \cite{cui2020towards} is proposed to increase the rank of the embeddings of all possible constituent labels.
The NML is defined as:
\begin{equation}
\begin{split}
\small
&\hat{C} = Embedding(C) \\
&\mathcal{L}_{NML} = - \frac{1}{N} {\| \hat{C}  \|}_{*}.
\end{split}
\end{equation}
where $C$ is the set of all possible constituent labels, $\hat{C}$ and $N$ are the embedding and length of $C$ respectively. ${\| \hat{C}  \|}_{*}$ is computed as:
\begin{equation}
\small
\label{eq:numclear}
{\| \hat{C}  \|}_{*} = \sum_{i} \sigma_i
\end{equation}
where  $\sigma_i$ is the $i$-th the singular value of $\hat{C}$.




\subsection{TTS with syntactic representations}
\label{ssec:metho_4_bnm}
The learnt syntactic representations are word related, which are upsampled to phoneme level and concatenated with phoneme embeddings.
Syntactic representation is copied to match the phoneme sequence length of current word.
Tacotron 2 \cite{shen2018natural} is employed to generate spectrogram from the concatenated syntactic representations and phoneme embeddings, and Griffin-Lim \cite{griffin1984signal} is further utilized to reconstruct the waveform.
The whole model is trained with a loss function which can be formulated as:
\begin{equation}
\small
\label{eq:bnm}
\mathcal{L} = \mathcal{L}_{TTS} + \lambda \mathcal{L}_{NML}
\end{equation}
where $\mathcal{L}_{TTS}$ is the loss function defined in Tacotron 2 and $\lambda$ is the loss weight for NML.


\section{Experiment}
\label{sec:experiment}


\subsection{Training setup}

We train models on public Chinese female corpus \cite{databaker}, which includes 10-hour professional speech and 10000 sentences. 
500 sentences are used for validation and other sentences are used for training.
We down-sample the speech to 16k Hz sampling frequency, 
The tacotron 2 part in our model is trained with vanilla setups except setting frequency to match our speech.
The learning rate is fixed to $ 10^{-3} $ and the loss weight of NML is 0.05. 

We train the WRF based TTS \cite{guo2019exploiting} as baseline approach.
The parser used in WRF \cite{levy2003harder} is replaced by state-of-the-art syntactic parsing model Benepar \cite{kitaev2018multilingual}.

We program all the models based on an open sourced Tensorflow implemention of Tacotron 2 \cite{mama_deepminds_2019}.
We train all the models for 50k iterations with a batch size of 16 on a NVIDIA 2080 Ti GPU.

\subsection{Subjective evaluation}
\label{ssec:exp_1_phythm}
We randomly select 30 sentences from Internet as test set, 5 of which are sentences with multiple different syntactic parse trees. 
Synthesized speeches are shifted in random order and rated by 20 native speakers on a scale from 1 to 5, from which a subjective mean opinion score (MOS) is calculated. 

As show in Table.\ref{table:mos}, the proposed system receives a MOS of $3.82$ while the baseline approach receives a MOS of $3.70$, with a comparable variance.
T-test reveals that our proposed approach significantly outperforms the baseline with a $p$ of $0.0079$.

We also conduct a ABX preference test between pairs of systems on the synthesized speech. 
The listeners are presented with the speeches synthesized by the baseline and proposed approaches in random order, and decide which one has the better prosody and naturalness.
As show in Table.\ref{table:mos2}, the proposed approach receives $45.8\%$ preference rate exceeding the baseline approach by $17\%$.

\begin{table}[!htb]
\renewcommand{\arraystretch}{1.3}
\caption{Comparision between baseline and the proposed method. MOS variances are given in brackets.}
\label{table:mos}
\centering
\begin{tabular}{c r r}
			\toprule
			 & WRF & Proposed\\ \midrule
            MOS & $3.70(0.05)$ & $3.82(0.06)$ \\ 
			\bottomrule
	\end{tabular}
\end{table}

\begin{table}[!htb]
\renewcommand{\arraystretch}{1.3}
\caption{ABX preference comparision between baseline and the proposed method. ABX-PR means preference rate of ABX test.}
\label{table:mos2}
\centering
\begin{tabular}{c r r r}
			\toprule
			 & WRF & Proposed & Neutral\\ \midrule
            ABX-PR & $28.8\%$ & $45.8\%$ & $25.4\%$ \\ 
			\bottomrule
	\end{tabular}
\end{table}


\begin{figure}[htb]
    \centering
    \includegraphics[width=.85\linewidth]{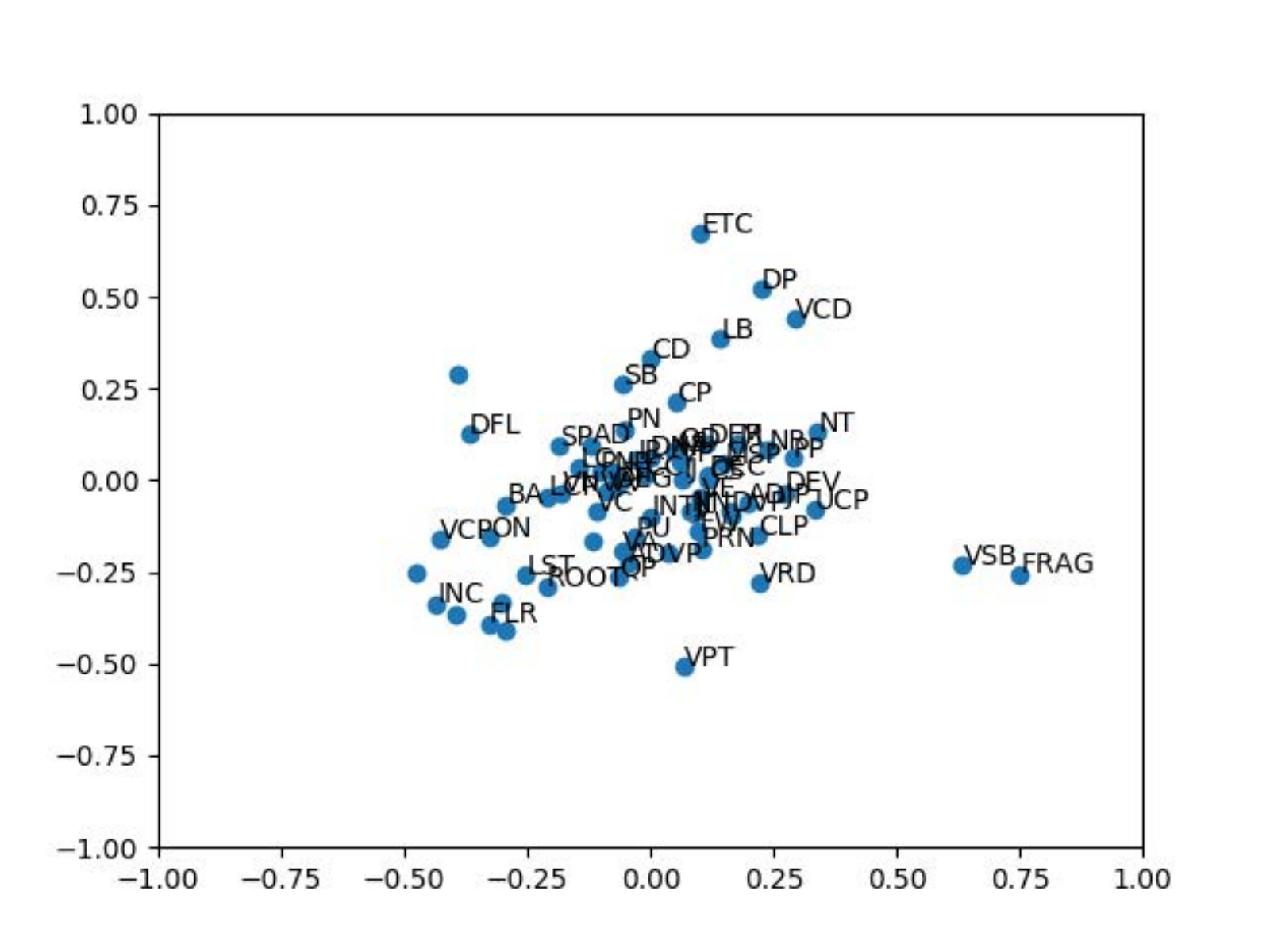}\\
    (a) Without NML
    \includegraphics[width=.85\linewidth]{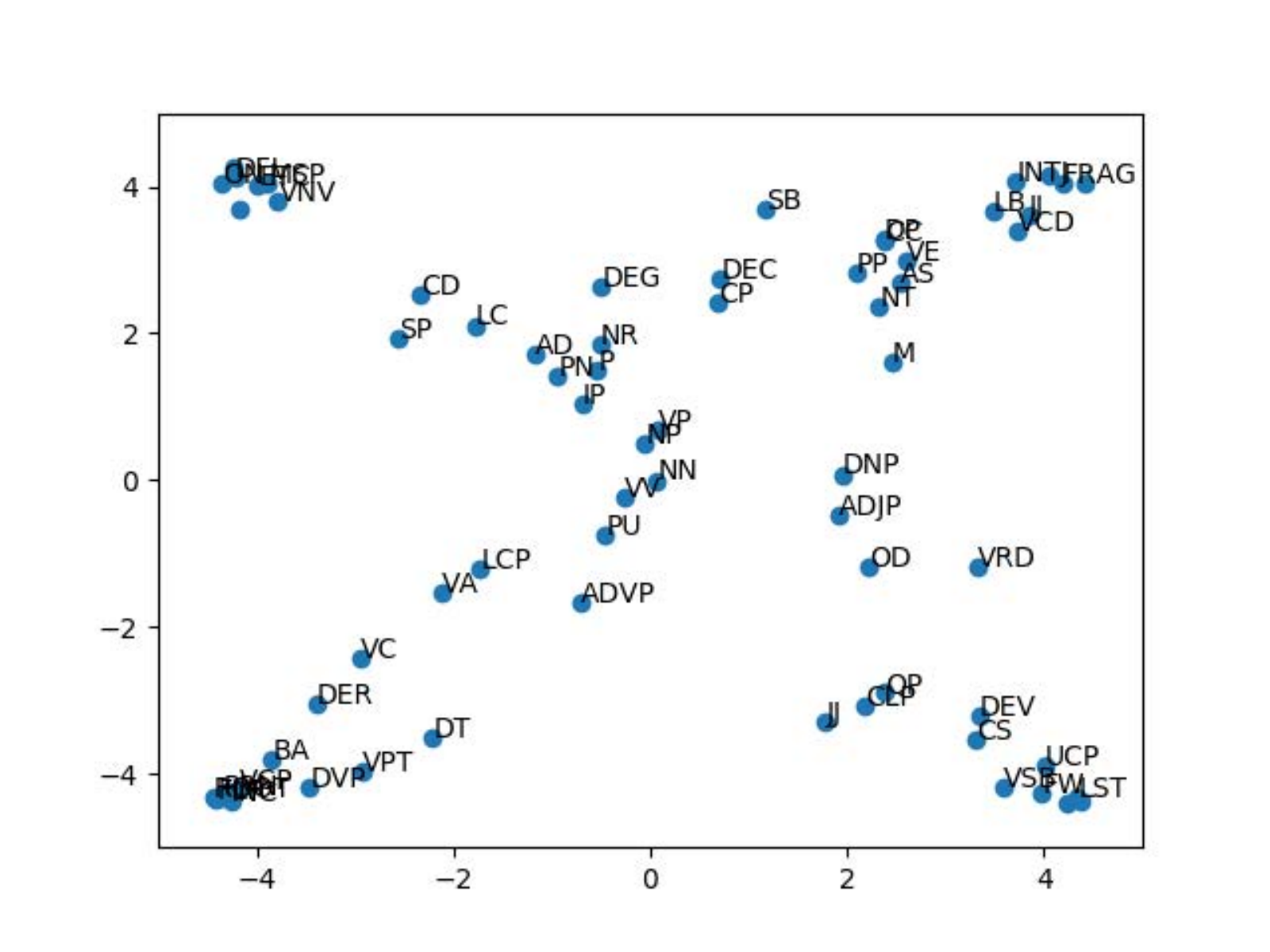}\\
    (b) With NML
      \caption{PCA results of constituent label embeddings on NML ablation study.}
  \label{fig:ablation}
\end{figure}

\subsection{Ablation study}
\label{ssec:exp_1_phythm}
To visualize the contribution of NML, we train our model without NML with the same settings.
Another ABX preference test is conducted on same test set and listeners.
The listeners are presented with the speeches synthesized by our models with and without NML, and decide which one has the better prosody and naturalness.
As show in Table.\ref{table:abx}, the approach with NML receives $58.3\%$ preference rate exceeding the approach without NML by $27\%$.

\begin{table}[!htb]
\renewcommand{\arraystretch}{1.3}
\caption{ABX preference result with or without NML.}
\label{table:abx}
\centering
\begin{tabular}{c r r r}
			\toprule
 			 & Proposed  & Proposed & Neutral\\
 			 & w/o NML & w/ NML & \\
 			 \midrule
             ABX-PR & $31.3\%$ & $58.3\%$ & $10.4\%$ \\ 
			\bottomrule
	\end{tabular}
\end{table}



We visualize the learnt embeddings of constituent labels with and without NML by principal components analysis (PCA).
As show in Fig.\ref{fig:ablation}, embeddings with NML are more scattered than the embeddings without NML, which demostrates the effectiveness of NML in improving discriminability and diversity.


\subsection{Analysis and discussion}
\label{ssec:exp_2_disam}

\begin{figure}[!htb]
	\centering
	\includegraphics[width=0.95\columnwidth]{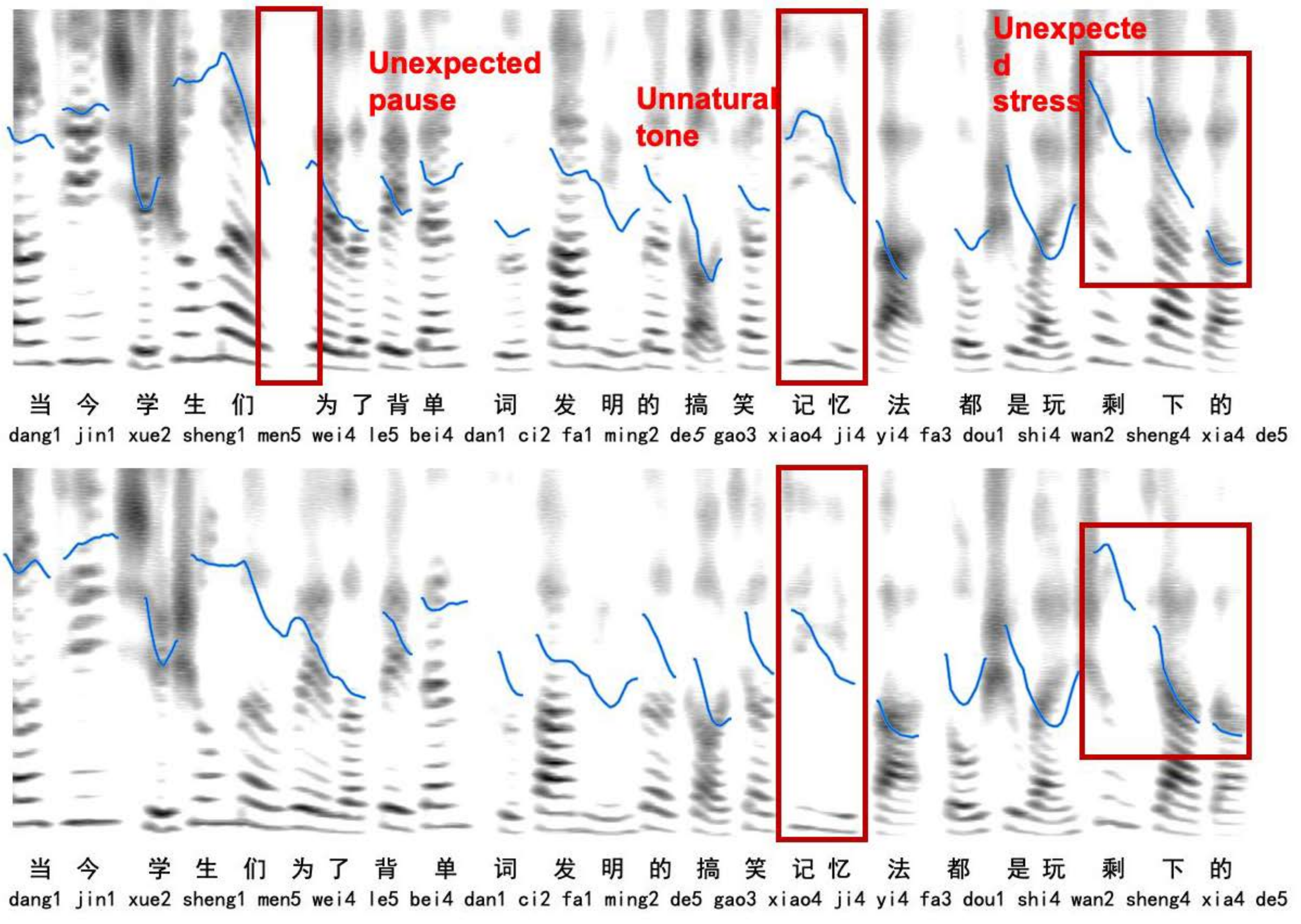}
	\caption{Mel spectrogram and pitch contour comparison on sentence ``At present, the funny memorization methods invented by students to memorize words are leftovers" between proposed model(below) and WRF(upper).}
	\label{fig:case}
\end{figure}



\begin{figure}[htb]
    \centering
    \includegraphics[width=.95\linewidth]{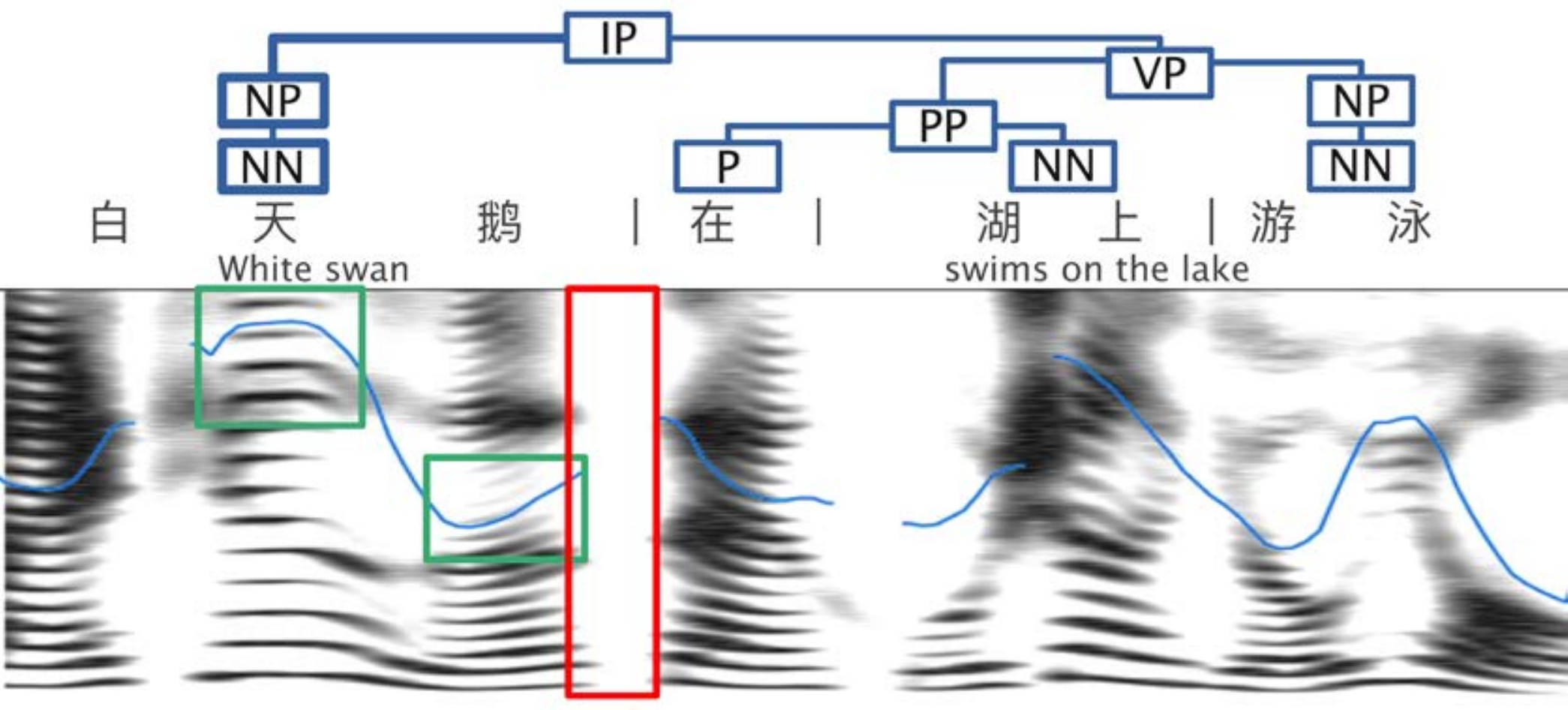}\\
    (a) White swan swims on the lake.
    \includegraphics[width=.95\linewidth]{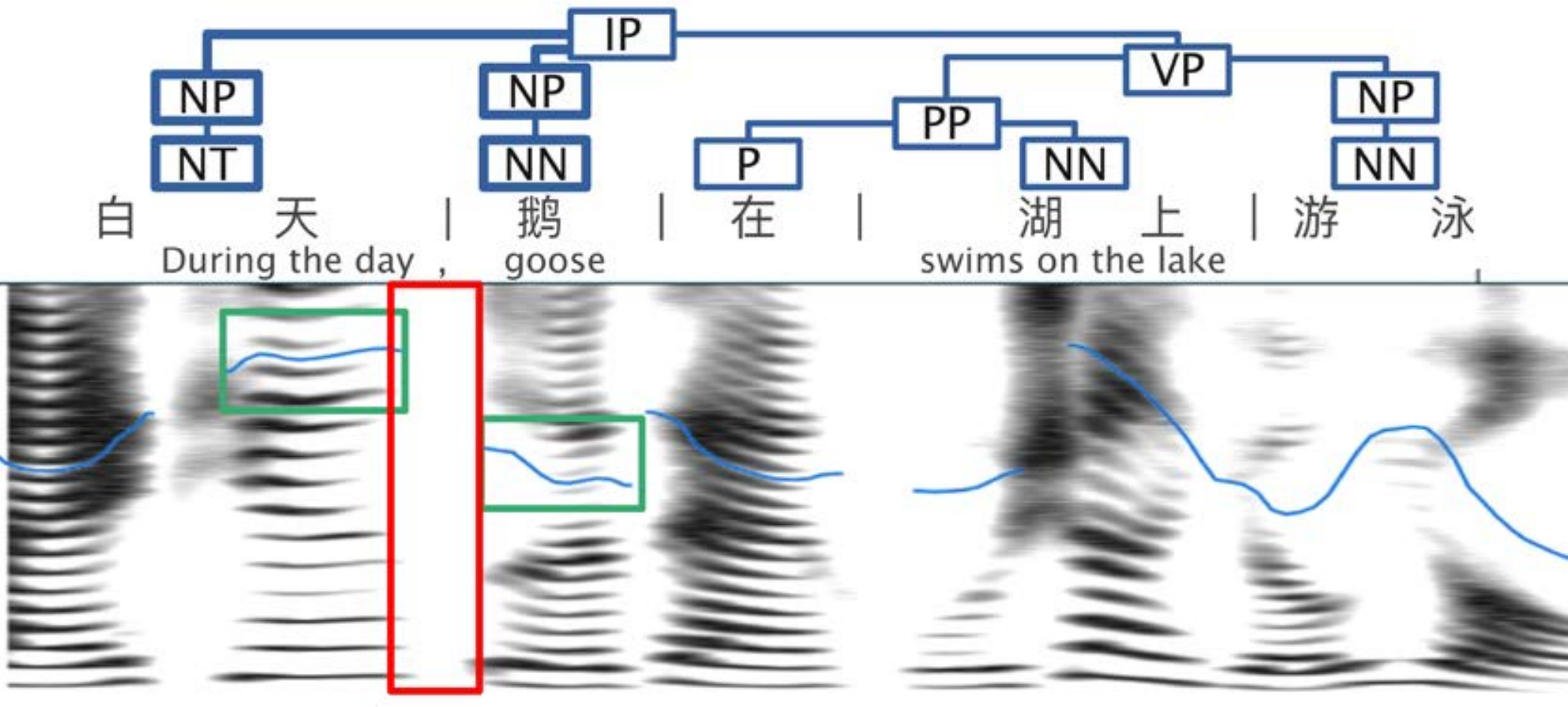}\\
    (b) During the day, goose swims on the lake.
      \caption{Prosodic differences of the sentence with multiple syntactic parse trees.}
    \label{fig:ambiguity_case}
\end{figure}

We further conduct case studies by comparing the spectrogram and pitch contours of the speeches generated from different methods, as shown in Fig.\ref{fig:case}.
The speech from baseline approach has an unexpected long pause between the 5-th character and 6-th character in such long sentence.

Besides, the word from the 16-th to 17-th character, ``ji4 yi4” that are both of fourth tone in Chinese, is synthesized with an unnatural up-and-down tone in WRF.
Instead, the spectrogram and pitch contour of the speech synthesized by our proposed method are more stable, indicating the stability of the proposed syntactic representations from bidirectional traversals.
For the last three characters, there is an unexpected stress (high pitch value) on 23-th character ``xia4" in WRF result, while proposed method shows a gradually decreasing pitch contour at the end of the sentence leading to higher naturalness. 
This is cause by the WRF features in baseline approach which consider a uni diridectional information in the syntactic parse tree.



With the proposed syntactic representation learning method, it is possible that a sentence with the same text but different syntactic parse trees might lead to synthesized speeches with different prosody expressions, which provides a possibility for prosody control of speech synthesis.  
To validate this, we conduct further experiments by inputting sentences with multiple different syntactic parse trees to the proposed model. 
One example is shown in Fig.\ref{fig:ambiguity_case}, from which the prosodic differences of the synthesized speeches can be clearly observed.
Upper structure regards the first three characters as a word and the sentence means ``White swan swims on the lake", while below treats these characters as two words and the sentence means ``During the day, goose swims on the lake". 
Although the graphemes and phonemes are same, the meanings of the sentences are different with each tree.
And the prosodic differences match the meanings respectively.
Also the prosody of last five characters in either synthesized speech are similar since their corresponding syntactic structure information is similar.

\section{Conclusion}
\label{sec:conclusion}

In this study, we investigate a syntactic representation learning method to automatically utilize the syntactic structure information for neural network based TTS.
Nuclear-norm maximization loss is introduced to enhance the discriminability and diversity of synthsized speech prosody.
Experimental results demonstrate the effectiveness of our proposed approach.
For sentences with multiple syntactic parse trees, prosodic difference can be clearly observed from the synthesized speeches.



\bibliographystyle{IEEEbib}
\bibliography{strings,refs}

\end{document}